%% file: main.tex
\documentclass[letterpaper, 10 pt, conference]{IEEEtran}
\IEEEoverridecommandlockouts
\usepackage{balance}
\usepackage{cite}
\usepackage{amsmath,amssymb,amsfonts,bbm,mathtools}
\usepackage{hyperref}

\usepackage[ruled, linesnumbered]{algorithm2e}
\SetAlCapNameFnt{\footnotesize}
\SetAlCapFnt{\footnotesize}

\usepackage{graphicx,subcaption}
\usepackage{textcomp}
\usepackage{xcolor}
\usepackage{comment}
\usepackage{multirow}
\let\labelindent\relax
\usepackage{enumitem}
\usepackage{lipsum}
\usepackage{arydshln}

\def\BibTeX{{\rm B\kern-.05em{\sc i\kern-.025em b}\kern-.08em
T\kern-.1667em\lower.7ex\hbox{E}\kern-.125emX}}

\DeclareMathOperator*{\argmin}{arg\,min}

\let\oldnl\nl
\newcommand{\nlnonumber}{\renewcommand{\nl}{\let\nl\oldnl}}

\begin{document}

\title{\LARGE \bf Reeling It In: Flexible Needle Pick Up via Thread Manipulation for Autonomous Suturing}

\author{
{Emma Huang$^{1}$,  Zih-Yun Chiu$^{2}$, Neelay Joglekar$^{3}$, Shanglei Liu$^{4}$, Michael C. Yip$^{1}$ \IEEEmembership{Senior Member, IEEE}}
\thanks{$^{1}$Department of Electrical and Computer Engineering, University of California San Diego, La Jolla, CA 92093 USA. {\tt\small\{emh002, yip\}@ucsd.edu}}
\thanks{$^{2}$Department of Computer Science, Johns Hopkins University, Baltimore, MD 21218 USA. {\tt\small zchiu@jhu.edu}}
\thanks{$^{3}$CMU Robotics Institute, Carnegie Mellon University, Pittsburgh, PA 15213 USA. {\tt\small njogleka@andrew.cmu.edu}}
\thanks{$^{4}$UC San Diego Health, La Jolla, CA, 92093 USA. {\tt\small s5liu@health.ucsd.edu}}
}
\maketitle

\begin{abstract}
Suture-needle pickup is necessary for autonomous suturing, as a needle can be unexpectedly dropped or strategically released to adjust the grasping configuration. 
Current methods for autonomous needle pickup typically guide a robot to move straight toward the needle and grasp it, limited to conditions where the needle is observable and directly approachable. 
In addition, grasping the needle lying on tissue can lead to the robot pinching nearby tissue or the needle jumping around due to its slippery surface, posing potential safety issues. 
This work proposes an autonomous framework that uses a suture thread as an assistive tool for indirect needle pickup, avoiding unnecessary tool-tissue contact and enabling pickup even when the needle is occluded or inaccessible. 
The framework spans the entire workflow, including thread and tissue reconstruction, safe grasp-point selection, stable thread lifting, and bimanual thread-following until securing needle grasping. 
The robot policies account for visual uncertainty to maximize robustness in real-world environments. 
We evaluate the proposed framework on a da Vinci Research Kit under various real-world conditions. 
The results demonstrate robust performance even with a challenging thread configuration or a non-approachable needle, closing the gap in applying autonomous robot policies to unstructured suturing environments.

\end{abstract}

\input{sections/introduction}

\input{sections/related_work}
\input{sections/methods}
\input{sections/results}
\input{sections/discussion_and_conclusion}

\bibliographystyle{IEEEtran}
\bibliography{ref}

\end{document}

%% file: sections/introduction.tex
\section{Introduction}
Autonomous robotic surgery has drawn increasing attention in recent years due to its potential to relieve surgeons of time-consuming, tedious tasks, particularly during prolonged operations~\cite{wah2025revolutionizing}. 
One such task suitable for automation is suturing, which involves constant manipulation of a suture needle and thread. 
Suturing occurs in conjunction with cavity exploration, organ mobilization, tissue dissection, and tool changes, requiring repeated placement and retrieval of the needle during the operation.
This step can be challenging due to the needle's thinness, small size, and slipperiness.

Prior studies have proposed methods for automating suture needle pickup that primarily guide the robot directly to the needle and grasp it~\cite{d2018automated,selvaggio2019haptic,ozguner2021visually,schwaner2021autonomous,bendikas2023learning,haworth2025suturebot}, with the assumption that the needle is observable and reachable.
However, the needle can be occluded or inapproachable when it is out of the camera view or hidden behind tissue or other instruments~\cite{medina2018needle}. 
In addition, direct needle pickup can cause the gripper to pinch the tissue too hard if the robot is not controlled carefully, potentially leading to tissue damage. 
Current methods have yet to address these challenges, requiring suturing environments to be highly structured.

To avoid unnecessary tool-tissue contact and expand the applicability of autonomous pickup to unreachable needles, an alternative method is to use the suture thread as an assistive tool for indirect needle pickup. 
The suture thread accounts for the majority of the suture length and can be deformed into many configurations, offering more degrees of freedom for manipulation than the rigid needle~\cite{shadrin2001handling}. 
Thread-assisted needle pickup is especially beneficial for needle occlusion or unreachable configurations: 
The robot can first grasp the free suture thread, then follow the thread's configuration to grasp the needle reliably~\cite{joglekar2025autonomous}. 
This alternative needle-pickup strategy reduces the risk of the robot pinching sensitive tissue and helps avoid thread entanglement, both of which are important for successful suturing~\cite{hari2025stitch}.

\begin{figure}[t!]
    \centering
    \includegraphics[width=0.99\linewidth]{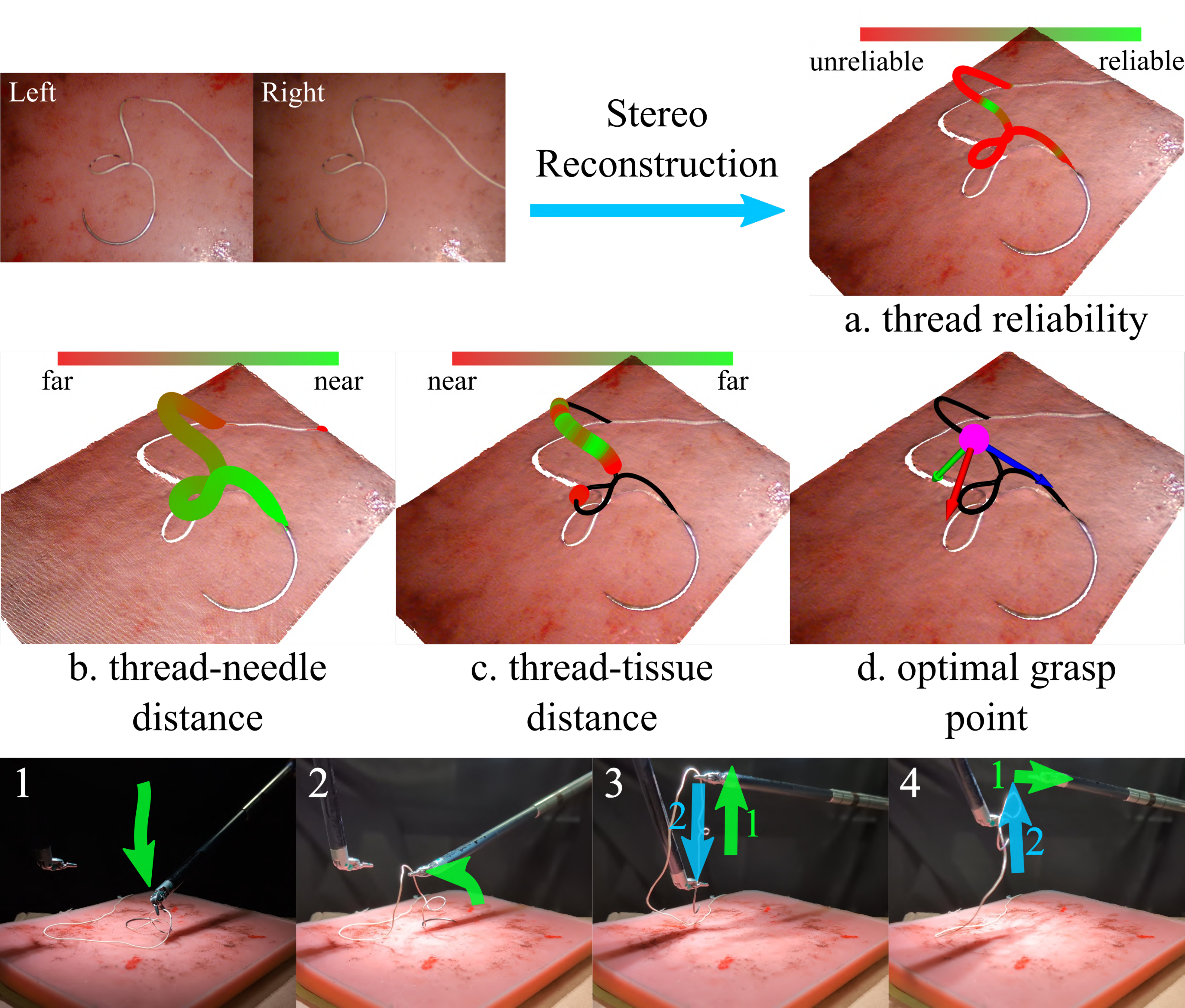}
    \caption{Our proposed suture-needle pickup framework. Our method leverages the suture thread configuration to identify the optimal thread-grasping point for safe needle lifting. Then, the robot follows a bi-manual thread traversing policy to stabilize and grasp the needle.}
    \label{fig:cover_image_v4}
    \vspace{-8mm}
\end{figure}

This work proposes a needle pickup framework that leverages suture thread for safe, reliable pickup and provides a complete pipeline from image observations to robot actions (Fig.~\ref{fig:cover_image_v4}). 
We extend prior methods~\cite{joglekar2025autonomous} to reconstruct the surgical environment and generate the optimal thread-grasping point by solving a reliability-aware optimization problem, minimizing the risk of tissue damage during grasping. 
The robot then completes a set of context-aware maneuvers to fish up and reel in the needle.

We evaluate our needle pickup framework on a da Vinci Research Kit (dVRK)~\cite{kazanzides2014open} under various real-world conditions.
The results demonstrate that the robot achieves over 95\% success rates when the needle and thread have minimal overlap and occlusions. 
When the thread is placed with loops and overlaps, which significantly complicates thread reconstruction, our method still achieves a 70\% success rate. 
In the most challenging scenarios, where the needle and thread are partially occluded, our method successfully grasps the needle for 65\% of the trials, demonstrating its robustness in real-world environments. 
In addition, we show that our reliability-aware grasping point selection and grasp planning lead to a 53.33\% improvement in successful thread grasping and less tissue pinching. 
Finally, our circular thread-lifting motion effectively eliminates needle dragging, improving safety during needle fishing. 
These experiments and analyses demonstrate that our method is an effective and safe alternative to needle pickup when directly approaching the needle is unsafe or impossible.

Our contributions are presented as follows:
\begin{itemize}
    \item A complete needle pickup framework that leverages the geometry and dynamic properties of suture thread.
    \item A reliability- and context-aware grasping policy robust to 3D reconstruction error and diverse configurations.
    \item The effectiveness of our method is demonstrated in various real-world scenarios, including challenging needle-thread configurations and occlusions.
\end{itemize}

%% file: sections/related_work.tex
\section{Related Work}

Suture needle manipulation is a fundamental task in autonomous suturing and has been addressed by various automation strategies. 
Most prior work assumes the needle is initially grasped by a robotic instrument and focuses on automating the subsequent suturing subtasks~\cite{sen2016automating,pedram2017autonomous,watanabe2017single,zhong2019dual,lu2020dual,mikada2020suturing,cao2020sewing,chiu2021bimanual,wilcox2022learning,hari2024stitch,hari2025stitch}. 
While these methods are effective for many suturing subtasks, they do not directly address scenarios in which the needle has been dropped and must be retrieved.

Another line of research focuses on developing policies for a robot to pick up the needle. 
\cite{d2018automated,ozguner2021visually} develops a visual servoing system for approaching and grasping the needle. 
\cite{selvaggio2019haptic} proposes a haptic-guided shared control method to pick up the needle while avoiding configuration singularities and joint limits. 
Recent studies have adopted learning-based methods, such as imitation learning and reinforcement learning, to learn a needle-pickup policy from data~\cite{schwaner2021autonomous,bendikas2023learning,haworth2025suturebot,ho2025surgirl}. 
These approaches are effective when the needle is mostly visible and approachable. 
This work further releases this assumption by proposing an indirect needle-pickup method that manipulates the suture thread.

Prior work on suture thread manipulation aims to grasp a manually selected point on a thread given its localization results~\cite{chow2013improved,chow2014knot}. 
Later approaches use deep learning to improve thread localization and incorporate a failure recovery strategy for thread grasping~\cite{lu2020learning_iros,lu2022grasp}. 
Recently, 3D thread reconstruction with reliability estimates has been proposed to provide uncertainty information about thread localization~\cite{joglekar2023suture,joglekar2025autonomous}. 
These reliability estimates are then used to guide thread grasping, resulting in higher success rates~\cite{joglekar2025autonomous}. 
Our work extends this method and proposes an uncertainty-aware thread-needle pickup strategy that accounts for contact between the thread and tissue to enable safe grasping-point selection, and considers the dynamics of the thread and needle to enable stable pickup.

%% file: sections/methods.tex
\section{Methods}
Our proposed suture-needle pickup framework includes the following steps: 
3D thread and tissue reconstruction, thread grasping-point selection, thread grasping and lifting, and bi-manual needle fishing to reel in the needle.
Algorithm~\ref{alg:needle-pickup} summarizes the complete process. 
\setlength{\textfloatsep}{0.1cm}
\begin{algorithm}[t!]
    \SetAlgoLined
    \caption{Needle Pickup with Bi-Manual Thread Following}
    \label{alg:needle-pickup}
    \footnotesize
    \KwIn{left image $\mathbf{I}_L$, right image $\mathbf{I}_R$, positions of gripper 1 and 2 $\mathbf{p}_{ee,1}$ and $\mathbf{p}_{ee,2}$, maximum workspace height $h_{workspace}$, the jaw's open width (half) $w$, uncertainty region of the thread configuration $\theta$, bi-manual grasping orientation $\mathbf{R}_{bi-grasp}$}
    \tcp{Obtain thread masks}
    $\mathbf{M}_L,\mathbf{M}_R \leftarrow segmentThread(\mathbf{I}_L, \mathbf{I}_R)$ \label{alg:equ-segment-thread}\\

    \tcp{Thread and tissue reconstruction, Sections~\ref{subsec:thread-reconstruction} and \ref{subsec:tissue-reconstruction}}
    $\mathcal{P}, \overline{\mathcal{K}}, \mathcal{R} \leftarrow threadReconstruction(\mathbf{I}_L, \mathbf{I}_R)$ \label{alg:equ-thread-reconstruction}\\
    $\mathbf{Q}_{filtered}, \mathbf{n} \leftarrow tissueReconstruction(\mathbf{I}_L, \mathbf{I}_R, \mathbf{M}_L,\mathbf{M}_R)$ \label{alg:equ-tissue-reconstruction}\\

    \tcp{Thread grasping, Sections~\ref{subsec:grasp-point-selection} and \ref{subsec:thread-grasping}}
    $\mathbf{p}_{g} \leftarrow threadGraspPointSelection(\mathcal{P}, \overline{\mathcal{K}}, \mathcal{R}, \mathbf{Q}_{filtered}, \mathbf{n})$

    $\mathbf{p}_{ee,1}, \mathbf{R}_{ee,1} \leftarrow threadGrasping(\mathcal{P}, \mathbf{p}_{g})$ \\

    \tcp{Calculate thread length between the grasping-point and the needle}
    $l_{thread} \leftarrow threadLength(\mathcal{P}, \mathbf{p}_{g})$
    
    \tcp{Circular thread pickup and thread lengthening, Sections~\ref{subsec:rotate-pick-up} and \ref{subsec:thread-lengthen}}
    $\mathbf{p}_{ee,1}, \mathbf{R}_{ee,1} \leftarrow threadPickupAndLengthen(\mathcal{P}, \mathbf{p}_{g}, h_{workspace})$

    \tcp{Bi-manual needle fishing, Section~\ref{subsec:bi-manual-fishing-motion}, Algorithm~\ref{alg:bi-manual-fishing}}
        $bimanualNeedleFishing(\mathbf{p}_{ee,1}, \mathbf{p}_{ee,2}, w, \theta, $\\
        \nlnonumber $\quad l_{thread}, h_{workspace}, \mathbf{n}, \mathbf{R}_{bi-grasp})$ \\
\end{algorithm}
\setlength{\floatsep}{0.1cm}

\subsection{Thread Reconstruction}
\label{subsec:thread-reconstruction}
We used the work of Joglekar et al.~\cite{joglekar2023suture,joglekar2025autonomous} to obtain 3D thread points and reconstruction reliability. 
The thread is first segmented from a pair of stereo images to generate thread segmentation masks, $\mathbf{M}_L$ and $\mathbf{M}_R$.
Stereo matching is done on the masks to calculate the disparity of thread pixels. 
The disparities are then re-projected into 3D space to form a set of unordered keypoints $\hat{\mathbf{k}}_1,\dots,\hat{\mathbf{k}}_N \in \mathbb{R}^3$. 
The keypoints will determine the locations of the 3D thread points.

We adopted a manual keypoint ordering method instead of the one used in~\cite{joglekar2023suture, joglekar2025autonomous} to handle overlaps and occlusions better. 
This ordering method is essential for successfully manipulating complex thread configurations.
The keypoints are ordered from the needle to the thread's tail as $\mathbf{k}_1,\dots,\mathbf{k}_N$. 

The ordered keypoints first define the local reliability region in the camera frame, ${region}(\mathbf{k}_i) \subset \mathbb{R}^3$ of the thread.
For each keypoint, a depth bound $\epsilon_{i, z}$ centered on $\mathbf{k}_i$
is defined with a least squares line $L_{\mathbf{k}_i}$ to form $\epsilon _{i,z} = 1.5 \Vert L_{i}(\mathbf{k}_{i, z}) - \mathbf{k}_{i,z}\Vert_2$. The depth bound is then split into a lower and upper bound, $\underline{\mathbf{k}}_i, \overline{\mathbf{k}}_i$ defined at each keypoint, to facilitate grasping-point selection in Section~\ref{subsec:grasp-point-selection}.
\begin{equation}
\begin{aligned}
    \underline{\mathbf{k}}_i & = \mathbf{k}_{i, z} - 1.5\| L_{\mathbf{k}_i}(\mathbf{k}_i) - \mathbf{k}_{i, z} \|_2\\
    \overline{\mathbf{k}}_i & = \mathbf{k}_{i, z} + 1.5\| L_{\mathbf{k}_i}(\mathbf{k}_i) - \mathbf{k}_{i, z} \|_2\\
\end{aligned}
\end{equation}
An additional pixel bound, parameterized by $\epsilon _{i,u}, \epsilon _{i,v}$, is defined in the image space to handle uncertainty due to segmentation. Combined with the depth bound, the local reliability region is defined as follows:
\begin{align} 
B(s_{i}) &\in {region}(\mathbf{k}_{i}) \iff \\ 
&\begin{vmatrix}\frac{1}{B_{z}(s_{i})} \pi(B(s_{i})) - \frac{1}{\mathbf{k}_{i,z}} \pi(\mathbf{k}_{i}) \end{vmatrix} \leq 
\begin{bmatrix}\epsilon _{i,u} \\ 
\epsilon _{i,v} \end{bmatrix}\\ 
& \begin{vmatrix}B_{z}(s_{i}) - \mathbf{k}_{i,z} \end{vmatrix} \leq \epsilon _{i,z},  
\end{align}
where $B$ is the 3D spline of the thread to be reconstructed, and $\pi: \mathbb{R}^3 \rightarrow \mathbb{R}^2$ is the pinhole camera model. 

A 3-D spline $B(s): \mathbb{R} \xrightarrow{} \mathbb{R}^3$ parameterized by $s \in [0,1]$ is then constructed using a constrained optimization with a smoothing loss following the method of~\cite{joglekar2025autonomous}. Additional explanation of the optimization is omitted for brevity, but can be read from~\cite{joglekar2025autonomous}. 
\begin{align} 
&\mathop{\mathrm{arg min}}_{B(s)} \mathcal {L}(B(s))\\ 
\text {s.t. } &\forall \mathbf{k}_{i} \,\exists s_{i} : B(s_{i}) \in {region}(\mathbf{k}_{i})
\end{align}
The spline, $B(s)$, is uniformly sampled at $N$ points to get $\mathcal{P}=\{\mathbf{p}_1,...,\mathbf{p}_N\}$. A reliability value $\mathcal{R}=\{r_1,...,r_N\}$ is assigned to each point by an inverse relation to the size of the nearby reliability regions. Each point is also paired with their lower bounds and upper bounds  $\underline{\mathbf{K}}=\{\underline{\mathbf{k}}_1,...,\underline{\mathbf{k}}_N\}$, and $\overline{\mathbf{K}}=\{\overline{\mathbf{k}}_1,...,\overline{\mathbf{k}}_N\}$. 
These values will determine the optimal grasping-point in Section~\ref{subsec:grasp-point-selection}. The reconstructed thread and corresponding reliability values can be seen in Fig.~\ref{fig:cover_image_v4}a.

\subsection{Tissue Reconstruction}
\label{subsec:tissue-reconstruction}
Stereo matching is used to create a point cloud representation of the tissue, 
$\mathcal{Q} 
= 
\left\{ \mathbf{q}_i \in \mathbb{R}^3 \right\}_{i=1}^{M}
$. 

We assume the normal of the surface $\mathbf{n}$ is parallel to the gravity $\mathbf{g}$, i.e., $| \mathbf{g} \cdot \mathbf{n} | \approx 1$. 
Given the pinhole camera model, $\pi: \mathbb{R}^3 \rightarrow \mathbb{R}^2$, and the thread mask, $\mathbf{M}_L$, only the points not overlapped with the thread are kept to remove inaccurate floating tissue points.
\begin{gather}
    \mathcal{Q}_{\text{non-thread}} =
    \left\{
    \mathbf{q}_i \in \mathcal{Q}
    \;\middle|\;
    \mathbf{M}_L\big(\pi(\mathbf{q}_i)\big) = 0
    \right\}    
\end{gather}
Additional tissue points within a radius $r \in \mathbb{R}$ from the thread are also filtered: 
\begin{align}
\mathcal{N}_r(\mathbf{q}_i) 
&=
\left\{
\mathbf{q} \in \mathcal{Q}_{\text{non-thread}} 
\;\middle|\;
\|\mathbf{q} - \mathbf{q}_i\|_2 \le r
\right\}
\\
\mathcal{Q}_{\text{filtered}} 
&=
\mathcal{Q}_{\text{non-thread}}\ \backslash\ \mathcal{N}_r(\mathbf{q}_i)
\end{align}

The normal vector of the tissue surface, $\mathbf{n}$, is obtained by averaging the normal vectors from all tissue points.
\begin{equation}
\begin{aligned}
\text{Given} \quad & \mathbf{n}_i^\top \mathbf{n}_j \ge 0 \quad \forall i \neq j \quad \text{and} \quad
\|\mathbf{n}_i\|_2 = 1, \\
& \mathbf{n} = \frac{1}{M}\sum_{i=1}^{M}\mathbf{n}_i
\end{aligned}
\end{equation}
$\mathcal{Q}_{\text{filtered}}$ and $\mathbf{n}$ provides the context for thread grasping and lifting from Sections~\ref{subsec:grasp-point-selection} to \ref{subsec:rotate-pick-up}. 
The reconstructed tissue can be seen in Fig.~\ref{fig:cover_image_v4}a-d.

\subsection{Thread Grasping-Point Selection} 
\label{subsec:grasp-point-selection}

\begin{figure}[t!]
    \centering
    \vspace{1.5mm}
    \includegraphics[width=0.99\linewidth]{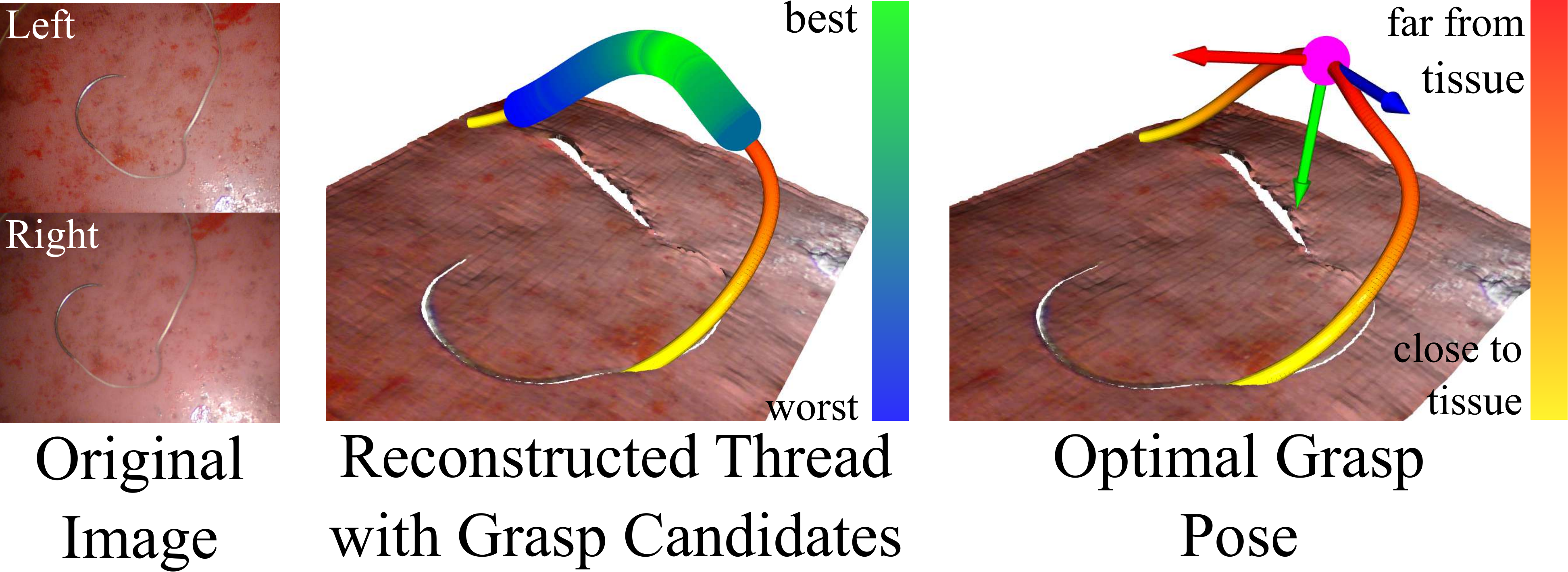}
    \caption{Stereo images are fed into the pipeline to reconstruct the thread and tissue point cloud. The optimal grasping point is selected based on the reconstruction reliability of the thread point, the distance to the needle, the distance to the tissue, and the steepness of the local thread segment. The candidates are shown ordered from green to blue in the center image, and the optimal grasping point is shown in pink in the right image. The coordinate frame represents the gripper's grasping orientation.
    }
    \label{fig:grasp_point_selection}
\end{figure}

The optimal grasping-point on the thread, $\mathbf{p}_{grasp}$, for needle pickup is the one closest to the needle while least likely to pinch the tissue during grasping. 
The optimal grasping point is found by solving the following constrained optimization problem. The constraints and variables are explained later.
\begin{align}
\argmin_{i \in \{1, \dots, N\}} \quad &
\alpha\ l(\mathbf{p}_i)-\beta\ d(\mathbf{p}_i, \mathcal{Q}_{\text{filtered}}) 
\label{equ:grasp-maximize}
\\
\text{s.t. } \quad & 
\overline{\mathbf{k}}_{i,z} + \delta_s < \mathbf{q}_{i,z} \quad \forall \mathbf{q}_i \in \mathcal{Q}_{\text{filtered}} \label{equ:bound-constraint}\\
&r_i > \delta_r \label{equ:reliability-constraint}\\
&\|\mathbf{p}_{i-1,z}-\mathbf{p}_{i,z}\|_2 < \delta_v \label{equ:velocity-constraint}
\end{align}

The grasping-point candidates are initialized to be all thread points, $\mathcal{C} = \mathcal{P}$.
A candidate is removed if it fails to satisfy the constraints in \eqref{equ:bound-constraint} to \eqref{equ:velocity-constraint}:
\begin{enumerate}
    \item Constraint \eqref{equ:bound-constraint}: The candidates must lie above the tissue. 
    To maximize the safety of thread grasping, we use the upper bound on a thread point's depth from the camera to encapsulate the maximum depth at which the point can be.  The candidates that satisfy this constraint are shown in Fig.~\ref{fig:cover_image_v4}c.
    An additional safety margin, $\delta_s \in \mathbb{R}$, is applied to ensure the thread point does not contact the tissue.
    \begin{align} \label{equ:bound-constraint-set}
        \mathbf{C}_T = \{\overline{K}(\mathbf{p}_i)_z + \delta_s < \mathbf{q}^*_{i,z}
        \;\mid\;
        \mathbf{p}_i \in \mathcal{P} \}
    \end{align}
    
    \item Constraint \eqref{equ:reliability-constraint}: The candidates' reliability is above a threshold, $\delta_r \in \mathbb{R}$. The reliability constraint is shown in Fig.~\ref{fig:cover_image_v4}a.
    This ensures that the grasping-point is within an acceptable error bound defined by its reliability.
    \begin{align}
        \mathbf{C}_R = \{R(\mathbf{p}_i) > \delta_r
        \;\mid\;
        \mathbf{p}_i \in \mathcal{P} \}
    \end{align}
    \item Constraint \eqref{equ:velocity-constraint}: 
    The local thread segment does not change much in depth. 
    This prevents selecting a grasping-point on a steep segment of thread that the gripper can easily miss.
    \begin{equation}
        \begin{aligned}
        \mathbf{C}_V = \{& | \mathbf{p}_{i-1,z} - \mathbf{p}_{i,z} | \leq \delta_v\ \cap\ 
        | \mathbf{p}_{i+1,z} - \mathbf{p}_{i,z} | \leq \delta_v \\
        & |\; 
        \mathbf{p}_i \in \mathcal{P} \}
        \end{aligned}
    \end{equation}
\end{enumerate}
The grasping-point candidates become $\mathcal{C} = \mathcal{C}_S\ \cap\ \mathcal{C}_R\ \cap\ \mathcal{C}_V$ shown in Fig.~\ref{fig:grasp_point_selection} as a gradient from green to blue.

With the k-nearest neighbor searching algorithm, we find the nearest point on the tissue, $\mathbf{q}^*_i$, for each thread point in $\mathcal{P}$ for distance $d(\mathbf{p}_i, \mathcal{Q}_{filtered})$.
\begin{gather}
d(\mathbf{p}_i, \mathcal{Q}_{\text{filtered}}) = \min_{\mathbf{q} \in \mathcal{Q}_{\text{filtered}}} \left\|
\mathbf{p}_i - \mathbf{q}
\right\|_2 \label{equ:distance_to_tissue}\\
\mathbf{q}_i^{*}
=
\arg\,min_{\mathbf{q} \in \mathcal{Q}_{\text{filtered}}}
\left\|
\mathbf{p}_i - \mathbf{q}
\right\|_2, \quad 
\mathcal{Q}^{*}
=
\left\{
\mathbf{q}_i^{*}
\right\}_{i=1}^{N}.
\label{equ:knn}
\end{gather}

The optimization objective minimizes the length from the thread point to the needle, $l(\mathbf{p}_i) = \sum_{\tau=1}^{i-1} \| \mathbf{p}_{\tau+1} - \mathbf{p}_{\tau} \|_2$ shown in Fig.~\ref{fig:cover_image_v4}b, and maximizes the distance of the thread point from the tissue~\eqref{equ:distance_to_tissue}, with each term weighted by the hyperparameters, $\alpha$ and $\beta$. The candidate in $\mathcal{C}$ with the smallest value from the optimization objective~\eqref{equ:grasp-maximize} is selected as the optimal thread grasping-point, $\mathbf{p}_{grasp}$, shown in Fig.~\ref{fig:cover_image_v4}d. and Fig.~\ref{fig:grasp_point_selection} marked in pink.

\subsection{Thread Grasping}
\label{subsec:thread-grasping}

\begin{figure}[t!]
    \centering
    \vspace{1.5mm}
    \includegraphics[width=0.99\linewidth]{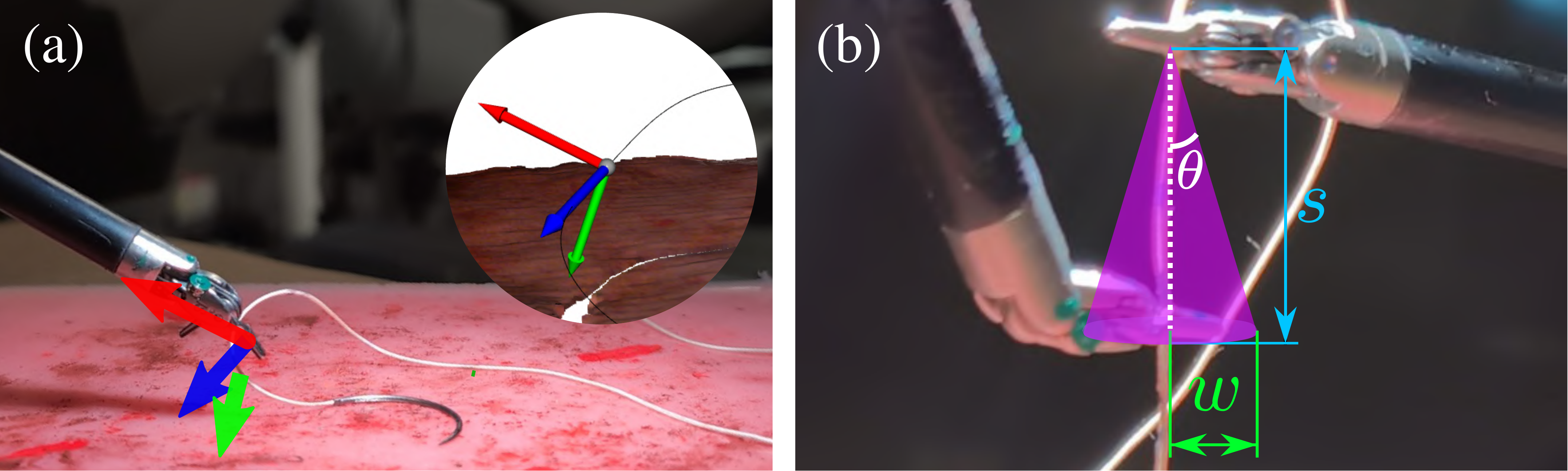}
    \caption{(a) Grasping orientation found using the thread and tissue reconstruction to generate the optimal suture grasping pose. (b) A cone modeling the uncertainty in the thread and needle configuration after lifting. The approximated flexural strength of the thread and the radius decided by the maximum gripper opening are used to find the step size $s$.}
    \label{fig:gripper-orientation-cone}
\end{figure}

In the following sections, the thread points, tissue reconstruction, and the optimal thread grasping point are used to grasp, lift, and pull the thread until reaching the needle.

With the point $\mathbf{p}_{grasp}$, we first define the gripper orientation that maximizes the grasp feasibility.
The gripper is oriented to approach from the outside of the thread's curvature, with its z-axis tangent to the thread to maximize the free region for grasping. 
The gripper orientation is shown in Fig.~\ref{fig:gripper-orientation-cone}a and defined as follows: 
\begin{gather}
\hat{\mathbf{z}}
=
\frac{
\mathbf{p}_{grasp} - \mathbf{p}_{prev}
}{
\left\|
\mathbf{p}_{grasp} - \mathbf{p}_{prev}
\right\|_2
}, \nonumber
\\
\hat{{\mathbf{x}}}
=
\mathbf{\hat{p}}_{grasp} \times \hat{\mathbf{z}}, \ \nonumber
\\
\hat{\mathbf{y}}
=
\hat{\mathbf{z}} \times \hat{\mathbf{x}}, \nonumber
\label{equ:thread-gripper-y-axis}
\\
\mathbf{R}_{grasp}
=
\begin{bmatrix}
\hat{\mathbf{x}} & \hat{\mathbf{y}} & \hat{\mathbf{z}}
\end{bmatrix}
\in SO(3),
\end{gather}
where $\mathbf{p}_{prev}$ is the adjacent thread point of $\mathbf{p}_{grasp}$ that is closer to the needle, and $\mathbf{\hat{p}}_{grasp}$ is the normalized $\mathbf{p}_{grasp}$. 

Note that since $\mathbf{p}_{grasp}$ satisfies constraint \eqref{equ:velocity-constraint}, taking the cross product of $\mathbf{p}_{grasp}$ and $\hat{\mathbf{z}}$ is feasible as they will not be aligned. The direction of the z-axis is chosen to always point along the thread towards the needle. This ensures that once the thread and needle are lifted, the gripper returns to an orientation that does not reach joint limits throughout the manipulation process. 
The direction of the y-axis points towards the tissue, which can be found by ensuring the dot product of the y-axis and the tissue normal is $< 0$.

During thread grasping, $\mathbf{p}_{grasp}$ is approached with a distance offset, $\delta_{offset} \in \mathbb{R}$, along the gripper's y-axis. 
The gripper first aligns with the orientation, $\mathbf{R}_{ee} = \mathbf{R}_{grasp}$, then translates toward $\mathbf{p}_{ee} = \mathbf{p}_{grasp} - \delta_{offset}\ \mathbf{\hat{y}}$. 

\subsection{Circular Thread Pick Up}
\label{subsec:rotate-pick-up}
\begin{figure}[t!]
    \centering
    \vspace{1.5mm}
    \includegraphics[width=0.85\linewidth]{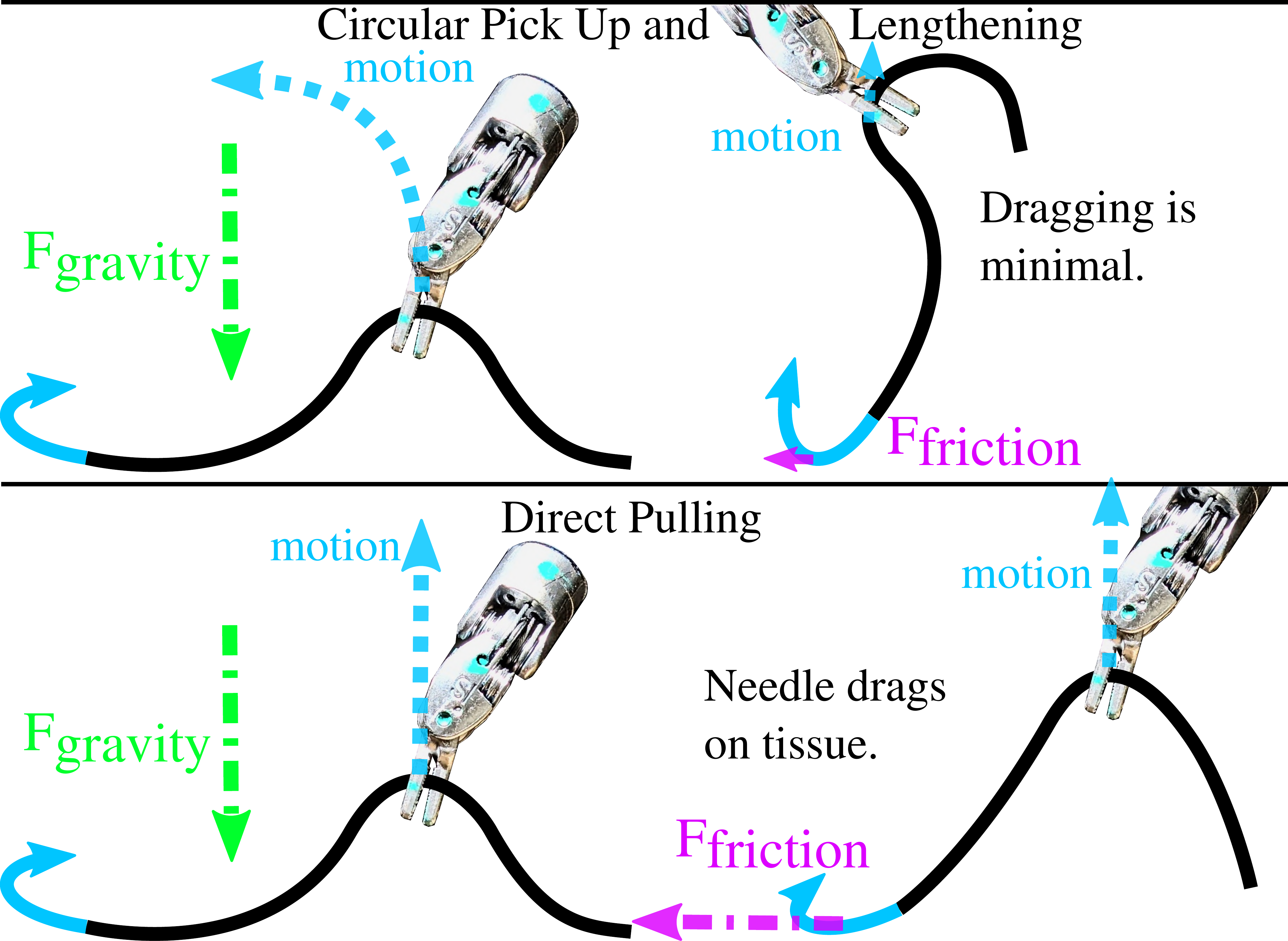}
    \caption{The needle can cause tissue damage when it drags on the tissue and embeds itself into the surface. By applying a rotation motion around the needle before lifting the thread along the direction of gravity, the dragging is largely reduced.}
    \label{fig:circular-thread-pickup}
\end{figure}

In this section, the thread and needle are lifted away from the tissue while maintaining minimal needle movement against the tissue. 
The robot grasps at the point, $\mathbf{p}_{grasp}$, and applies a circular motion around the needle tail point, approximately at $\mathbf{p}_1$, to lift the thread away from the tissue until the thread aligns with the tissue normal vector, $\mathbf{n}$.
Specifically, the rotation of the circular motion, $\Delta\mathbf{R}_{circular} \in SO(3)$, is found by aligning $\mathbf{p}_{grasp} - \mathbf{p}_1$ with $\mathbf{n}$: 

\begin{gather}
\mathbf{v} 
= 
\mathbf{p}_{grasp} - \mathbf{p}_1, \quad
\hat{\mathbf{v}}
=
\frac{\mathbf{v}}{\|\mathbf{v}\|_2}
\\
\phi
=
\arccos\!\left(
\hat{\mathbf{v}}^\top
\mathbf{n}
\right)
,\quad
\hat{\mathbf{a}}
=
\frac{
\hat{\mathbf{v}} \times \mathbf{n}
}{
\left\|
\hat{\mathbf{v}} \times \mathbf{n}
\right\|_2
}
\\
\Delta \mathbf{R}_{circular} = \text{AxisAngle2Matrix}(\hat{\mathbf{a}}, \phi),
\end{gather}
where $\text{AxisAngle2Matrix}(\cdot, \cdot)$ transforms an axis-angle rotation to a rotation matrix. 
The gripper orientation is then set to $\mathbf{R}_{ee} = \Delta\mathbf{R}_{circular} \mathbf{R}_{ee}$, and its position to $\mathbf{p}_{ee} = \mathbf{p}_{ee} + \|\mathbf{v}\|_2\ \mathbf{n}$. 
The robot movement is shown in Fig.~\ref{fig:circular-thread-pickup}.
The thread aligns with the gravity direction after being picked up in circular motion, as we assume $\mathbf{n}$ is parallel to the gravity. 

This circular motion minimizes the likelihood that the needle will hook or drag on the tissue during lifting, which can damage important organs.
Fig.~\ref{fig:cover_image_v4}(2). shows this motion with a green arrow, and Fig.~\ref{fig:circular-thread-pickup} compares our circular movement with a direct lifting strategy and shows how our strategy reduces the needle dragging against the tissue. 

\subsection{Thread Lengthening} \label{subsec:thread-lengthen}
The thread is straightened by lifting in the direction of $\mathbf{n}$. 
If the thread is longer than the height of the surgical workspace such that it cannot be fully extended, it is lifted to the maximum height $h_{workspace} \in \mathbb{R}$ possible.
The gripper is translated toward $\mathbf{p}_{ee} = \mathbf{p}_{ee} + d_l \mathbf{n},\ \mathbf{p}_{ee,z} = \min(\mathbf{p}_{ee,z}, h_{workspace})$, where $d_l \in \mathbb{R}$ is the distance that the gripper moves along $\mathbf{n}$. 

Note that the thread has been aligned with the gravity direction before lengthening, and the gripper movement is also parallel to it. 
This way, the thread dynamics is largely stabilized during lifting, as no significant force is applied from other directions. 
The stable thread dynamics further prevents the needle from contacting nearby tissues and enhances manipulation safety. This robot movement is shown in Fig.~\ref{fig:circular-thread-pickup} and Fig.~\ref{fig:cover_image_v4}(3). indicated with a green arrow.

\subsection{Bi-manual Needle Fishing}
\label{subsec:bi-manual-fishing-motion}

\begin{algorithm}[t!]
    \caption{Bi-Manual Needle Fishing}
    \label{alg:bi-manual-fishing}
    \vspace{0.5mm}
    \footnotesize
    \SetAlgoLined
    \KwIn{holding gripper's position $\mathbf{p}_{hold}$, grasping gripper's position $\mathbf{p}_{grasp}$, the jaw's open width (half) $w$, uncertainty region of the thread configuration $\theta$, thread length $l_{thread} = \| \hat{\mathbf{v}} \|_2$, maximum workspace height $h_{workspace}$, tissue normal vector $\mathbf{n}$, bi-manual grasping orientation $\mathbf{R}_{bi-grasp}$}
    
    \tcp{Determine the step size}

    $s = \frac{w}{\arctan{\theta}}$

    \uIf{$l_{thread} > h_{workspace}$}{
        \tcp{Generate fishing motions}
    
        $N_f \leftarrow \frac{l_{thread}}{s}$
            
        $\mathbf{goals} = \{ \mathbf{p}_{hold} - i \cdot s\mathbf{n}\ \text{ for } i \text{ in } \{1, \dots, N_f\}\}$
    }
    \Else{
        \tcp{Generate one motion to the needle}
        $N_f \leftarrow 1$
        
        $\mathbf{goals} = \{\mathbf{p}_{hold} - l_{thread}\mathbf{n}\}$
    }
    \tcp{Execute the motions toward the goals}
    \For{$i = 1,\dots, N_f$}{
        Move the grasping gripper toward $(\mathbf{goals}[i], \mathbf{R}_{bi-grasp})$. \\
        The grasping gripper closes the jaw. \\
        The holding gripper opens the jaw to release the thread and retracts. \\
        The grasping gripper lifts the thread and needle upward to the original height of the holding gripper. \\
        Switch the role of the grasping and holding grippers. 
     }
\end{algorithm}

After the thread is lifted and safely elevated from the tissue surface, we begin a bi-manual fishing motion to capture the needle. 
Once lifted, the thread configuration becomes difficult to predict because of the needle's weight distribution and the thread's flexural strength, which prevents the thread from being fully straightened by gravity. 
A thread with higher flexural strength retains its previous shape, resulting in random loops and bends that make it difficult to localize. 

To predict the possible states of the thread and needle after they are lifted, we model the uncertainty region of the thread as a cone parameterized by its potential deviation from the gravity direction, $\theta \in \mathbb{R}$, as shown in Fig.~\ref{fig:gripper-orientation-cone}b.
$\theta$ will determine the step size of bi-manual fishing motions. 

Two grippers will take turns grasping the thread until the needle is reached. 
Due to uncertainty in the thread configuration, each gripper will approach the thread with its gripper fully open. 
The distance from each finger to the thread is denoted as $w \in \mathbb{R}$.
The step size of bi-manual motions is determined by $s = \frac{w}{\arctan{\theta}}$. 
$w$ and $s$ are depicted in Fig.~\ref{fig:gripper-orientation-cone}. 

Algorithm~\eqref{alg:bi-manual-fishing} details the process of bi-manual needle fishing. 
Multiple fishing motions are used when the thread is too long to be lifted fully at once. 
Otherwise, the needle is directly reached in one fishing motion. This motion is traversed in Fig.~\ref{fig:cover_image_v4}-3 to 4.

%% file: sections/results.tex
\section{Experimental Results}
Our needle pickup method is evaluated in real-world experiments on a dVRK~\cite{kazanzides2014open}. 
The success rates of our full pipeline, Algorithm~\ref{alg:needle-pickup}, are reported across four setups: \textbf{Easy}, \textbf{Medium}, \textbf{Hard}, and \textbf{Occlusion} (Fig.~\ref{fig:easy-medium-hard-setup} and \ref{fig:occlusion-setup}). 
The importance of the thread reliability measures and the thread lifting method for mitigating tissue pinching and needle dragging is further assessed through ablation studies. 

A 36 mm 1/2-circle suture needle with a 45 cm thread was used in the experiments. 
We set the values of $\alpha$ and $\beta$ defined in equation~\eqref{equ:grasp-maximize} to 0.7 and 1, respectively, giving the distance from the thread points to the tissue a higher weight than the distance from the thread points to the needle. 
The safety margin $\delta_s$ defined in equation~\eqref{equ:bound-constraint-set} is set to 1 mm. The reliability threshold $\delta_r$ is set to $0.4$ in equation~\eqref{equ:reliability-constraint}. The velocity threshold $\delta_v$ is set to $0.6$ in equation~\eqref{equ:velocity-constraint}.
The thread grasping-point offset, $\delta_{offset}$, is set to 20 mm.
The step size $s$ for bi-manual fishing in Algorithm~\ref{alg:bi-manual-fishing} is set to 15 mm when $l_{thread} > h_{workspace}$. 

\subsection{Experimental Setups}
\begin{table*}[t!]
    \vspace{2mm}
    \caption{Success rates of the full pipeline under different setups}
    \centering
    \vspace{0.5mm}
    \begin{tabular}{c|cc:cc:cc:cc}
    Results & Easy Grasp    & Easy Pickup   & Medium Grasp  & Medium Pickup & Hard Grasp    & Hard Pickup   & Occlusion Grasp   & Occlusion Pickup  \\ \hline
    Success & 19            & 19            & 19            & 19            & 19            & 14            & 18                & 13                \\ 
    Fail    & 1             & 1             & 1             & 1             & 1             & 6             & 2                 & 7	                \\ \hline \hline
    Percent & 95\%          & 95\% & 95\%          & 95\% & 95\% & 70\%          & 90\%     & 65\%              \\ 
    \end{tabular}
    \label{tab:pipeline-results}
\end{table*}

\begin{figure}[t!]
    \vspace{1.5mm}
    \centering
    \includegraphics[width=0.99\linewidth]{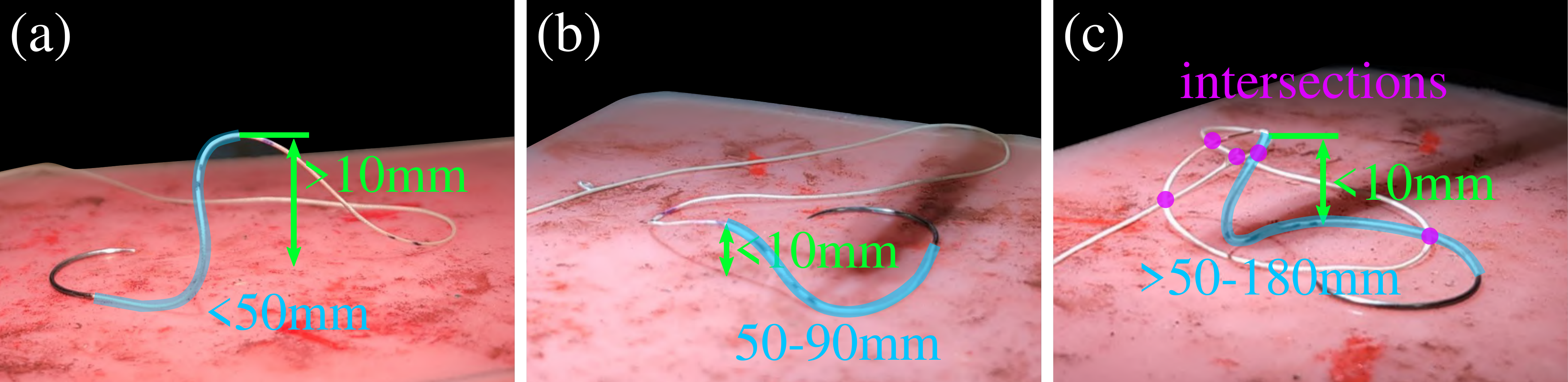}
    \caption{From left to right: \textbf{Easy}, \textbf{Medium}, and \textbf{Hard} setups, showing the differences from simplest setups to complex configurations.}
    \label{fig:easy-medium-hard-setup}
\end{figure}

\begin{figure}[t!]
    \centering
    \includegraphics[width=0.8\linewidth]{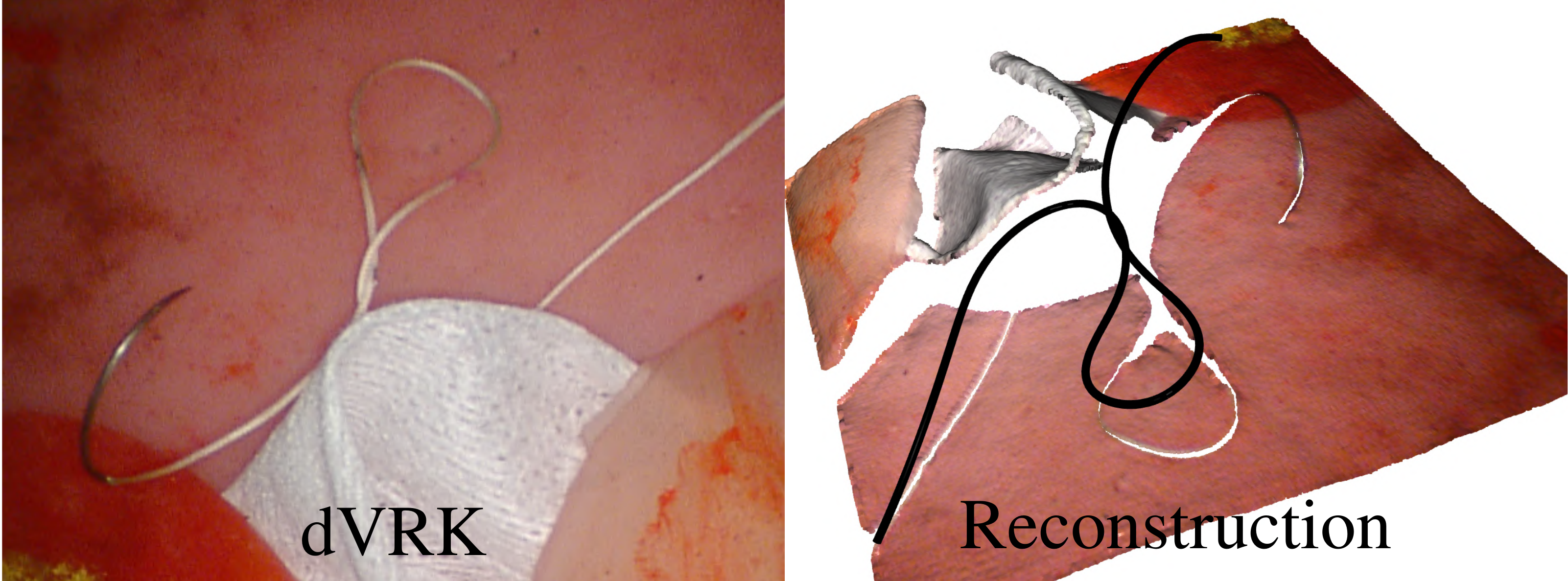}
    \caption{Occlusion setup from the dVRK endoscopic camera and a reconstruction result showing limited information in occluded regions.}
    \label{fig:occlusion-setup}
\end{figure}

The following four setups are used (Fig.~\ref{fig:easy-medium-hard-setup} and \ref{fig:occlusion-setup}): 
\begin{description}[style=unboxed,leftmargin=0cm]
    \item[Easy:] The thread and needle are placed with no overlap and minimal occlusions, the highest point is elevated at least 10~mm from the tissue, and the thread length is $\leq$ 50 mm. 
    \item[Medium:] The thread and needle are placed with no overlap and minimal occlusions, the highest point is elevated less than 10 mm from the tissue, and the thread length is within the range of 50 mm to 90 mm.
    \item[Hard:] The thread is placed with one or more loops/overlaps and minimal occlusions, the highest point is elevated less than 10 mm from the tissue, and the thread length is within the range of 50 mm to 180 mm.    
    \item[Occlusion:] The thread and needle are partially occluded with gauze and additional silicon phantoms to imitate occlusion due to gauze placement during laparoscopic surgery. The thread is randomly arranged with overlaps and elevated portions, and its length is set between 50 mm and 180 mm.
\end{description}

\begin{figure}[t!]
    \centering
    \includegraphics[width=0.99\linewidth]{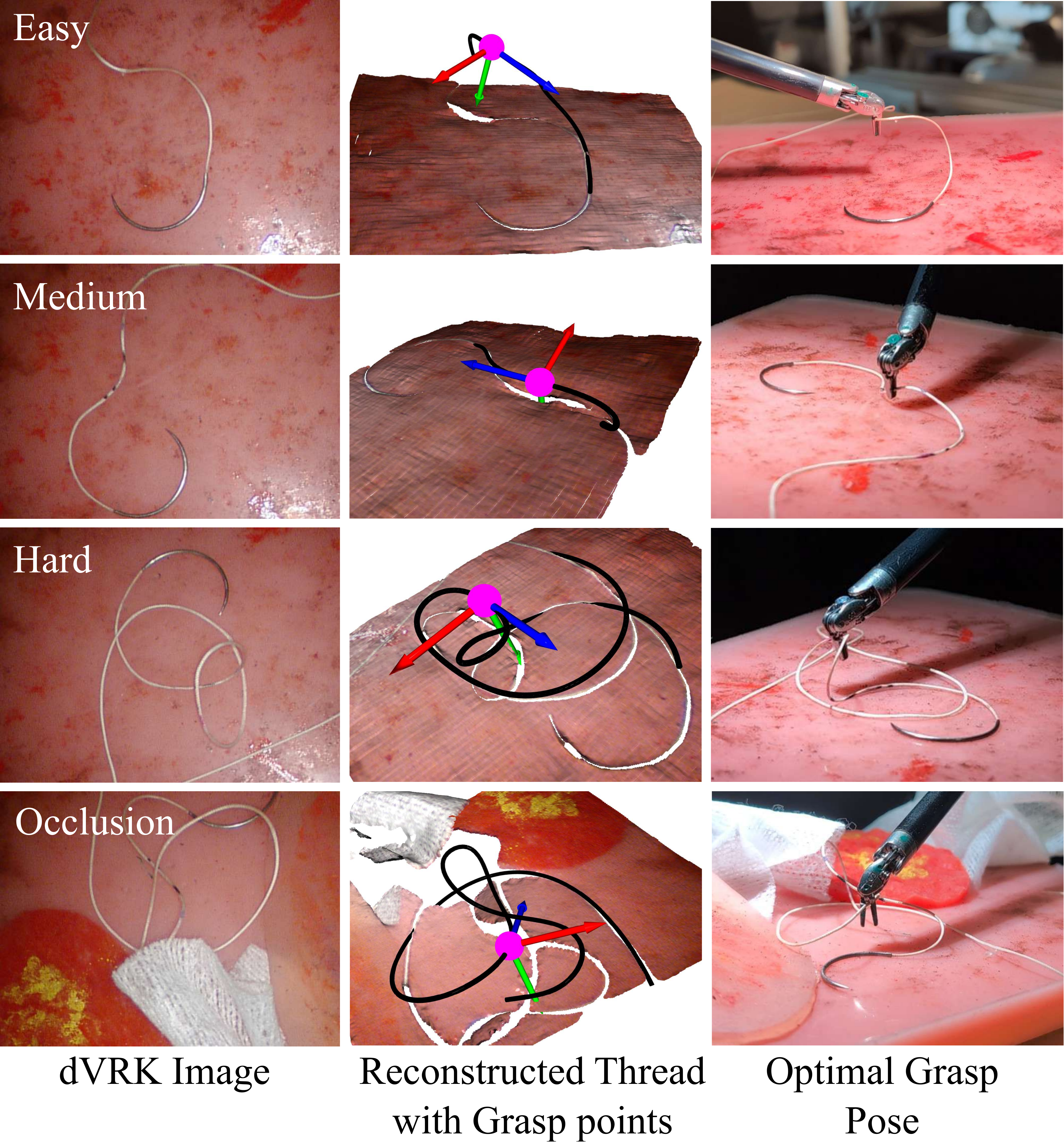}
    \caption{The optimal grasping-point is generated from the dVRK images (left column) and shown as the purple sphere in the middle column, with a coordinate frame representing the gripper orientation. The grasp is executed as shown in the right-most column.}
    \label{fig:easy-medium-hard}
\end{figure}
Examples of the thread and tissue reconstruction are shown in Fig.~\ref{fig:easy-medium-hard}. 
The middle column shows the optimal grasping point and the coordinate frame that represents the gripper's orientation.
The rightmost column shows side views of the setups to visualize the thread configurations better. 
In general, the optimal grasping point is selected on segments of thread that are vertical in the dVRK images. 
This is because their 2D to 3D reconstruction is more accurate as there is a bigger disparity between the left and right images. 

\begin{figure*}[t!]
    \centering
    \vspace{1.5mm}
    \includegraphics[width=0.99\linewidth]{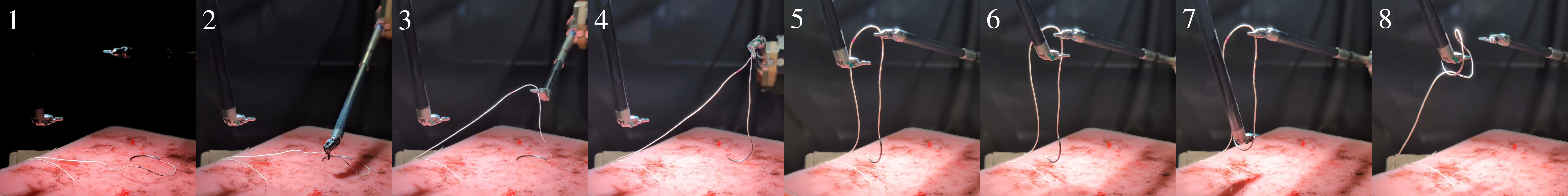}
    \caption{Manipulation pipeline: 1) The thread is set up with a small elevated portion. 2) The thread is grasped with the optimal grasping point chosen at the tip of the elevated portion. 3) A pivot around the needle is done to grasp the thread upright. 4) The thread is lengthened by pulling the thread upwards. The needle is lifted from the tissue surface. 5) The thread and needle are centered in front of the camera. 6) The second gripper approaches the thread and grasps it with its jaw. 7) The gripper moves along the thread to the needle. 8) The needle is grasped and lifted to show completion of the pipeline.}
    \label{fig:manipulation-pipeline}
    \vspace{-3mm}
\end{figure*}

Table~\ref{tab:pipeline-results} shows the success rates of the full pipeline. 
A Grasp trial is successful if the gripper grasps the thread at the optimal grasping point. 
A Pickup trial is successful if the gripper picks up the thread and reaches within 10 mm of the needle. 
Fig.~\ref{fig:manipulation-pipeline} shows an example of a robot performing thread grasping, circular pick up, lengthening, and bi-manual fishing to pick up the needle. 

Our method achieved a 95\% success rate under \textbf{Easy} and \textbf{Medium} setups. Each had one failed trial due to poorly reconstructed threads that led to missed grasps. 

Our method achieved a success rate of 70\% under the \textbf{Hard} setup, but it achieved an identical success rate of 95\% for thread grasping, demonstrating the robustness of the grasping-point selection method even with complex overlapping thread arrangements. 
However, there were more failure cases during bi-manual fishing, including the following: 
(1) Extra slack of the thread that allowed the needle to be anchored away from the gripper, pulling the thread out of the predicted uncertainty cone during bi-manual fishing. 
(2) The addition of loops and overlaps in the thread added uncertainty to the thread and tissue reconstruction near self-intersections, preventing the generation of grasping-points. 
(3) Overlapping portions of the thread were grasped together, causing the thread to hold its tangled shape and lie entirely outside the predicted uncertainty cone. 

The \textbf{Occlusion} setup achieved a comparable thread grasping success rate of 90\%. 
However, similar to the \textbf{Hard} setup, the success rate of bi-manual fishing is lower. The experiments revealed several notable failure cases: 
(1) The proximity of the gauze to most portions of the thread prevented generating grasping points. 
(2) The thread caught on the gauze during bi-manual fishing, pulling the thread out of the predicted uncertainty cone. 
(3) Multiple portions of the thread were grasped, causing the thread to loop over the gripper and remain outside of the uncertainty cone.

These failure cases suggest that improved reliability of thread and tissue reconstruction is necessary to reduce the safety margin for grasping-point selection and increase the number of grasping-point candidates. 
Improvements in the bi-manual fishing pipeline are also necessary to increase the accuracy of thread handoff when extra slack causes the thread to lie outside the predicted uncertainty cone.

\subsection{Ablation Studies}
\begin{table}[t!]
    \vspace{-2mm}
    \caption{Ablation Studies Success Rates}
    \centering
    \begin{tabular}{c|c:cc}
    Results             & No Pinching       & Reliability   & No Reliability    \\ \hline 
    Success             & 80                & 15            & 7                 \\
    Fail                & 0                 & 0             & 8                 \\ \hline\hline
    Percent             & 100.00\%          & 100.00\%      & 46.66\%           \\
    
    \end{tabular}
    \label{tab:ablation-results}
\end{table}
\subsubsection{Tissue Pinching}
The safety of the manipulation is evaluated by the number of times the gripper pinched and damaged the tissue. 
The results are recorded in Table~\ref{tab:ablation-results} for each trial where a success counts as the thread grasped without visible deformation of the tissue. 

In all of the Easy, Medium, Hard, and Occlusion trials, no pinching of the tissue occurred during manipulation. 
The results in Table~\ref{tab:ablation-results} show that the grasping-point selection strategy in Section~\ref{subsec:grasp-point-selection} ensures safe thread grasping using information about the thread reconstruction reliability and thread to tissue proximity. 
While the circular pick up motion, lengthening, and bi-manual fishing in Sections~\ref{subsec:rotate-pick-up}, \ref{subsec:thread-lengthen}, and \ref{subsec:bi-manual-fishing-motion} minimizes the risks of tool and tissue contact by working in the available space above the tissue.

\subsubsection{Thread Reliability Effects}

\begin{figure}[t!]
    \centering
    \includegraphics[width=0.9\linewidth]{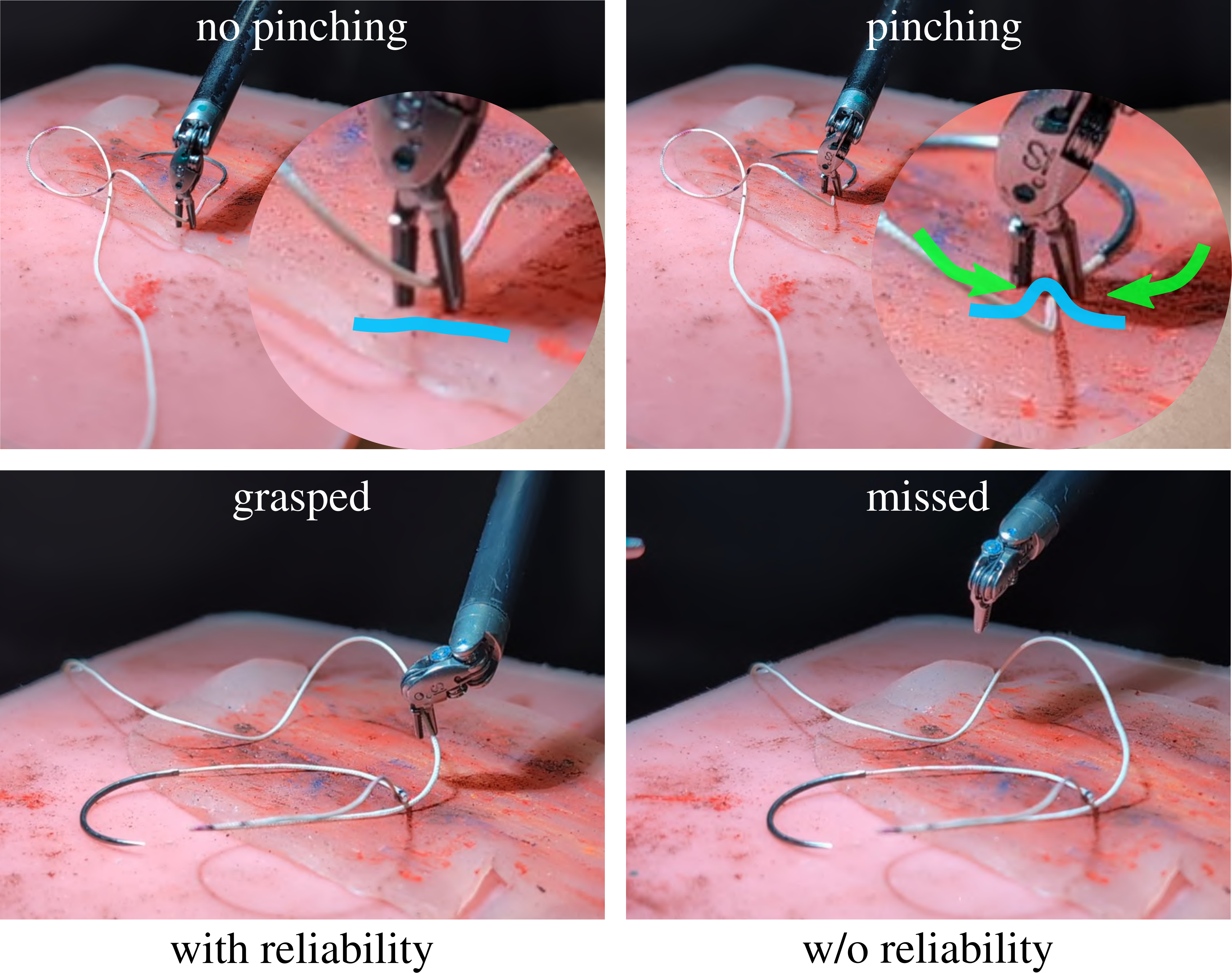}
    \caption{Comparison of the thread grasping motion shows that the gripper is less likely to cause tissue damage and miss the thread when the optimal grasping point is selected using the reliability metric.}
    \label{fig:reliability-pinching-missing}
\end{figure}

The importance of using the thread reliability metric for grasping-point selection was evaluated by running pairs of trials, with and without the metric, on the same thread configuration. 
The selected grasping-points were executed on the dVRK to assess thread grasping success and tissue damage. 

The results in Table~\ref{tab:ablation-results} showed that 8 out of 15 trials yielded successful grasps when the thread reliability metric was used, while grasping failed without the metric. 
In one trial, the grasping-point selected without the metric resulted in tissue pinching, which could have caused tissue tearing during manipulation. 
In the remaining failed trials, the gripper completely missed the thread due to thread reconstruction inaccuracies, incorrectly selecting a grasping-point far from the tissue and thread.
Fig.~\ref{fig:reliability-pinching-missing} shows the comparison of grasping-point selection with and without the thread reliability metric. 
In the seven successful trials without the thread reliability metric, both methods selected the same grasping-point along the thread and achieved successful grasps. 
No trials exhibited grasp failure when the reliability metric was used compared to when it was not. 
This ablation study demonstrates that incorporating the reliability metric into grasping-point selection improves grasp success by 53.33\% without causing additional failure cases. 

\subsubsection{Circular Pickup Effects}
\begin{figure}[t!]
    \centering
    \includegraphics[width=0.9\linewidth]{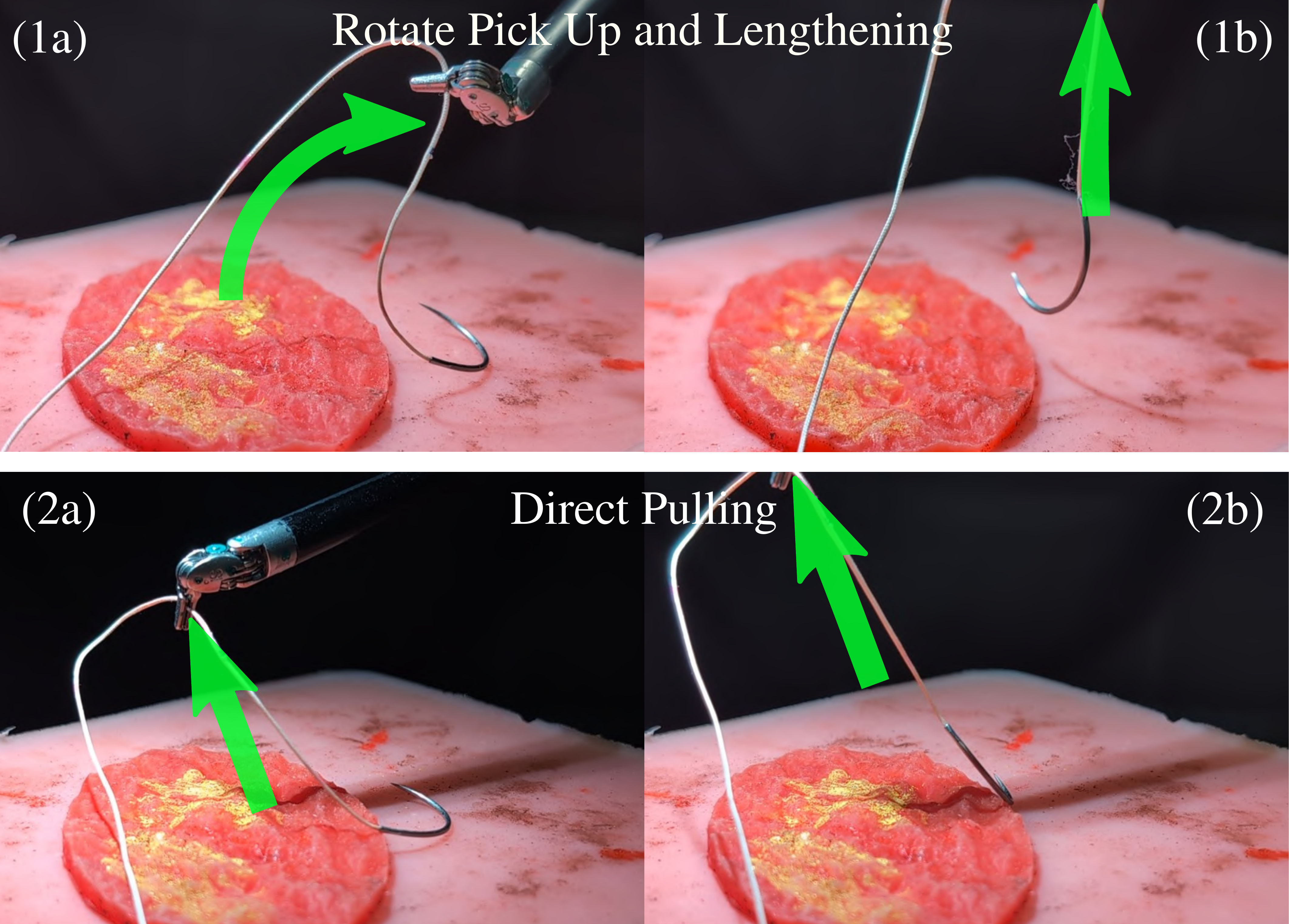}
    \caption{Frames 1a and 1b show minimal dragging of the needle on the tissue with our circular pick up and lengthening. Frames 2a and 2b show needle dragging on the tissue as the thread is lifted using a direct pulling method.}
    \label{fig:swinging-comparison_v2}
\end{figure}

The importance of the circular pickup and thread lengthening motions proposed in Sections~\ref{subsec:rotate-pick-up} and \ref{subsec:thread-lengthen} were evaluated in comparison to a direct lifting method. 
Minimal needle-to-tissue contact was observed with the circular pickup and thread lengthening method, compared with the direct lifting method, in which the needle was shown to drag across the tissue.
This comparison is demonstrated in Fig.~\ref{fig:swinging-comparison_v2}.

%% file: sections/discussion_and_conclusion.tex
\section{Discussion and Conclusion}
In this work, we propose a framework for autonomous suture-needle pickup that uses the thread to safely reel the needle in. 
Our framework integrates environment reconstruction, thread grasp-point selection, and manipulation into a complete pipeline. 

Experimental evaluations on the dVRK demonstrated that the proposed pipeline performed reliably across a diverse set of configurations - achieving 95\% success rates in \textbf{Easy} and \textbf{Medium} setups, while maintaining high grasp success even under challenging thread arrangements and occlusions. 
Failures in the \textbf{Hard} and \textbf{Occlusion} setups were primarily caused by high uncertainty in the reconstructions and increased slack that caused the thread to fall outside the predicted region during bi-manual fishing.

Ablation studies further validate our proposed grasp-point selection and circular pickup strategies. 
Using the thread reliability metric significantly improved grasping accuracy, while the circular pickup strategy effectively reduced needle dragging against the tissue compared to direct lifting, improving the safety of manipulation. 
Across all trials, the grasp-point selection policy successfully prevented tissue pinching, underscoring the importance of reliability-aware grasping and context-aware manipulation.

Our work provides a new strategy for needle pickup that leverages thread geometries, which can be directly followed by needle regrasping~\cite{lu2020dual,chiu2021bimanual,wilcox2022learning} to prepare for suturing. 
Future work will focus on improving thread and needle reconstruction under heavy occlusions and in complex surgical environments, and on incorporating real-time feedback to improve the accuracy of thread and needle manipulation. 
These improvements will better handle unpredictable thread deformations during bi-manual fishing. 
They will also be applied to tasks such as suture tightening, thread disentanglement, and suture throwing, which require real-time thread localization to be completed reliably. 
